\documentclass[sn-mathphys,Numbered]{sn-jnl}

\usepackage{graphicx}%
\usepackage{multirow}%
\usepackage{amsmath,amssymb,amsfonts}%
\usepackage{amsthm}%
\usepackage{mathrsfs}%
\usepackage[title]{appendix}%
\usepackage{xcolor}%
\usepackage{textcomp}%
\usepackage{manyfoot}%
\usepackage{booktabs}%
\usepackage{algorithm}%
\usepackage{algorithmicx}%
\usepackage{algpseudocode}%
\usepackage{listings}%
\usepackage{array}
\usepackage{makecell}
\usepackage{adjustbox}

\begin{document}

\title[Article Title]{A Rapid Review of Clustering Algorithms}

\author*[1]{\fnm{Hui} \sur{Yin}}\email{huiyin@swin.edu.au}

\author[1]{\fnm{Amir} \sur{Aryani}}\email{aaryani@swin.edu.au}


\author[1]{\fnm{Stephen} \sur{Petrie}}\email{spetrie@swin.edu.au}


\author[2]{\fnm{Aishwarya} \sur{Nambissan}}\email{aishwarya.nambissan96@gmail.com}

\author[1]{\fnm{Aland} \sur{ Astudillo}}\email{aastudillocontreras@swin.edu.au}

\author[2]{\fnm{Shengyuan} \sur{Cao}}\email{shengyuan.cao@anu.edu.au}


\affil[1]{
\orgname{Swinburne University of Technology}, 
\state{Victoria}, 
\country{Australia}}

\affil[2]{
\orgname{Australian National University}, 
\state{ Canberra}, 
\country{Australia}}


\abstract{Clustering algorithms aim to organize data into groups or clusters based on the inherent patterns and similarities within the data. They play an important role in today's life, such as in marketing and e-commerce, healthcare, data organization and analysis, and social media. Numerous clustering algorithms exist, with ongoing developments introducing new ones. Each algorithm possesses its own set of strengths and weaknesses, and as of now, there is no universally applicable algorithm for all tasks.
In this work, we analyzed existing clustering algorithms and classify mainstream algorithms across five different dimensions: underlying principles and characteristics, data point assignment to clusters, dataset capacity, predefined cluster numbers and application area. 
This classification facilitates researchers in understanding clustering algorithms from various perspectives and helps them identify algorithms suitable for solving specific tasks.
Finally, we discussed the current trends and potential future directions in clustering algorithms. 
We also identified and discussed open challenges and unresolved issues in the field.}


\maketitle

\section{Introduction}\label{sec_Introduction}

With the rapid transformation of digital technology, access to information has become a more prevalent aspect of daily life than ever before. Businesses and individuals now have access to vast information, encompassing billions of documents, videos, and audio files across the internet. 
Most businesses operate in an interconnected web of data linked to multiple information sources, capable of providing access to substantial data points. 
However, while our computing capabilities are rapidly growing, the sheer volume of data often challenges our capacity to transform disconnected data into useful information, practical knowledge, and actionable insights. One established solution is to leverage machine learning, particularly clustering methods.
Clustering algorithms are machine learning algorithms that seek to group similar data points based on specific criteria, thereby revealing natural structures or patterns within a dataset. The primary purpose is to divide the data into subgroups, or clusters, where items within the same group are more comparable to those in other sets. Clustering methods contribute to advancements in various fields, such as information retrieval, recommendation systems, and topic discovery.

Clustering algorithms are ubiquitous in daily life. They are used for spam email classification, recommendation systems, customer segmentation for targeted marketing, image processing for organizing images based on visual similarities, and more. Clustering algorithms can cluster text-based data and are also applicable to audio, video, and images. Audio clustering algorithms use acoustic features to group files and can be used for tasks like genre identification. Clustering video data facilitates tasks like recommendation and summarization by organizing content by visual or thematic similarities. In image analysis, clustering is essential for segmentation and content-based retrieval tasks. Clustering algorithms also act as a means of detecting anomalies, whether in network traffic, financial transactions, or medical records. Their versatility underscores their importance in extracting patterns and insights from a wide range of data.

Based on the nature of the learning process and the availability of labeled data, clustering algorithms are primarily categorized into two types:
\begin{itemize}
    \item Semi-supervised Learning: This category provides a training dataset for each data sample associated with a known cluster label. The algorithm observes the patterns in the training dataset and then learns to assign new data points to clusters. Such as Constrained K-Means~\cite{wagstaff2001constrained}, Semi-Supervised Fuzzy C-Means (SSFCM)~\cite{benkhalifa1999text}.
    \item Unsupervised Learning:
    In this category, no labeled dataset is provided. The algorithm identifies patterns and structures within the data and then groups similar data samples based on inherent similarities in the features without prior knowledge of the groupings. 
    Usually, the total number of clusters in the entire dataset is unknown. Examples of algorithms in this category include K-Means~\cite{Lloyd1982,Forgy1965}, Density-Based Spatial Clustering of Applications with Noise (DBSCAN)~\cite{ester1996density}, and Fuzzy C-Means (FCM)~\cite{bezdek1984fcm}. 
    Several methods can be employed to determine a suitable value, such as the elbow method, silhouette score, gap statistics, as discussed in Section~\ref{sec_PredifinedClusterNumbers}.
\end{itemize}

As new applications emerge, there is a growing demand for clustering algorithms that can effectively handle different types of data and scenarios.
At the same time, with the rise of big data and complex data structures, such as high-dimensional, heterogeneous data and large-scale datasets, adaptable clustering algorithms are needed to handle these different types of data effectively.
Therefore, clustering methodologies constantly evolve, so it becomes crucial to comprehensively survey and review existing literature to understand the latest developments. 
Several notable reviews have contributed significantly to this effort.
Rui Xu and Wunsch, D.~\cite{1427769} looked at clustering techniques for computer science, machine learning, and statistics datasets. They demonstrated how to use these algorithms on a few benchmark datasets: the traveling salesman issue, and bioinformatics—a relatively new topic garnering much attention. They also discussed cluster validation, closeness measures, and several related subjects.
Xu and Tian~\cite{xu2015comprehensive} conducted a comprehensive survey of clustering algorithms in 2015, covering a diverse range of techniques. The authors categorized these algorithms based on their underlying principles, characteristics, and applications. The survey includes a detailed and comprehensive comparison of all discussed algorithms.
Ezugwu et al.,~\cite{ezugwu2022comprehensive} provided an up-to-date, methodical, and thorough analysis of both conventional and cutting-edge clustering algorithms for various domains from a more practical viewpoint. It covered the application of clustering to various fields, such as big data, artificial intelligence, and robotics. The review also focused on the remarkable role that clustering plays in a variety of disciplines including education, marketing, medicine, biology, and bioinformatics.  
The three works mentioned above are the most comprehensive and highly cited research on clustering algorithms since 2000. Published in 2005, 2015, and 2022, these works have made significant contributions to the field. 

In addition to these seminal works, other notable pieces mostly focus on specific classifications or applications within clustering.
Bora et al.,~\cite{DBLP:journals/corr/BoraG14} conducted a comparative study between the fuzzy and hard clustering algorithms. The focus is on assessing and contrasting the performance of these clustering algorithms, providing insights into their strengths and limitations.
Sisodia et al.,~\cite{sisodia2012clustering} explored diverse clustering algorithms in the field of data mining, emphasizing fundamental aspects such as clustering basics, requirements, classification, challenges, and the application domains of these algorithms.
A comprehensive comparative analysis of 9 well-known clustering algorithms is provided by Rodriguez et al.,~\cite{rodriguez2019clustering}. 
The authors evaluated the performance and characteristics of these algorithms through a systematic evaluation, offering insights into their strengths and limitations. 
This study contributes to understanding clustering methods and facilitates informed choices based on specific application requirements.

In this work, we conduct a comprehensive summary of the existing clustering algorithm literature and classify it from four different perspectives to help users identify algorithms suitable for their specific tasks efficiently. Furthermore, we discuss an overview of the current research status and highlight future clustering technology trends.

\section{Method}\label{sec2}
In this section, we describe the review methodology, detailing the keywords used to collect publications and how they were screened for inclusion.
We defined a set of keywords related to our topic, such as ``clustering", ``clustering algorithm", ``clustering method", ``consensus clustering", ``clustering technique". 
After creating the keyword list, we conducted search across three reputable academic databases, including Google Scholar, arXiv, and Scopus. We employed Boolean operators (AND, OR) and truncation to refine search queries.
We chose Google Scholar because it includes papers that have not yet been formally published, such as preprints. We observed that some preprint publications received many citations at an early stage due to their significant contributions to the research field. However, because some journals require lengthy processing times before official publication, these publications may remain in preprint status for a long time.

We filtered the gathered publications based on the following criteria: publications written in English, published within the last five years, and focusing on novel clustering techniques.
Additionally, we removed duplicate papers not directly related to clustering algorithms to ensure that the remaining content comprises algorithm introductions, articles on algorithm technology improvement, and application articles. 
Next, we reviewed the titles, abstracts, and keywords, further screening the publications to narrow down the selection.
Once we concentrated on the full text of the selected publications, our aim was to identify the underlying principles of the algorithms, the algorithms used in their applications, the experimental procedures, and key findings.
We considered aspects such as experimental design, sample size, and statistical methodologies to assess the dependability of the results.
Finally, the results (presented in the next section) synthesize the patterns and trends found in the literature. 

\section{Results}\label{sec3}
\begin{figure}[ht]
    \centering
    \includegraphics[width=1.0\linewidth]{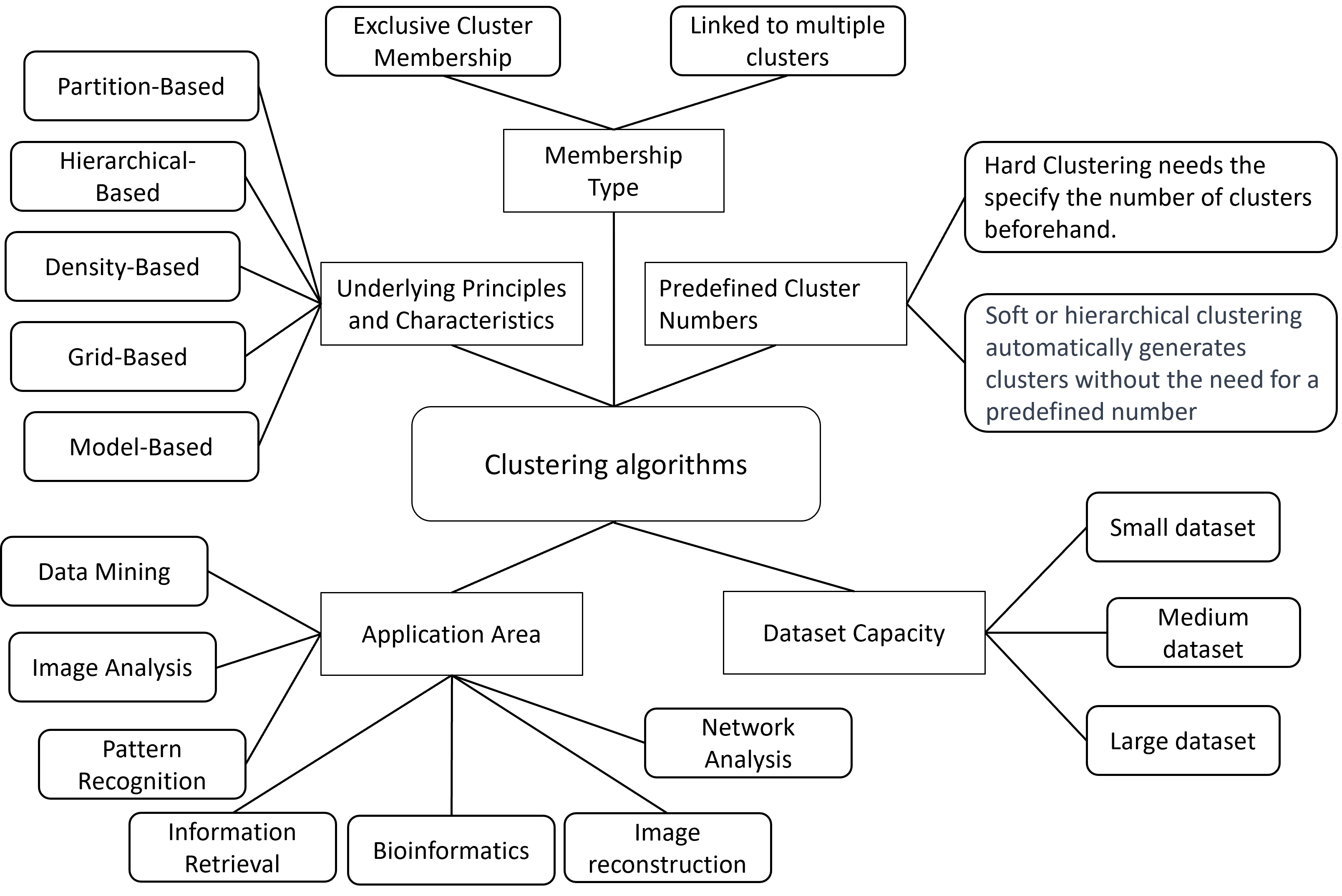}
    \caption{Structure of the clustering algorithm classification, covering five dimensions.}
    \label{fig_Structure}
\end{figure}

In this section, we analyze the fundamental characteristics and approaches of clustering algorithms and classify algorithms based on these principles, which is also currently the most recognized classification method. Subsequently, we classify the algorithms from different dimensions, such as the algorithm's capability to handle different dataset sizes, data point assignment to clusters, the requirement to predefine the number of clusters, and application area. This classification aims to guide users in choosing a suitable algorithm according to the specific clustering tasks.
The structure of the algorithm classification system is visually presented in Figure~\ref{fig_Structure}.

\subsection{Algorithm Classification by Underlying Principles and Characteristics}
Based on the fundamental characteristics and approaches to grouping data, clustering algorithms can be grouped into the following five distinct subsets~\cite{zhou2022comprehensive,sajana2016survey,berkhin2006survey} (summarised in Table~\ref{tab_Principles}): 
\begin{itemize}
    \item Partition-Based Clustering (e.g. Figure~\ref{fig:Partition-Based}): Partition-based clustering algorithms partition data into non-overlapping clusters. The basic idea of these clustering algorithms is to consider the center of data points as the center of the corresponding cluster, which generally has low time complexity and high computing efficiency. However, they are not well-suited for non-convex data, are sensitive to outliers, can be easily drawn into local optima, and require a predefined number of clusters, which can impact the performance of clustering results. 
    \begin{figure}[ht]
        \centering
        \includegraphics[width=1\linewidth]{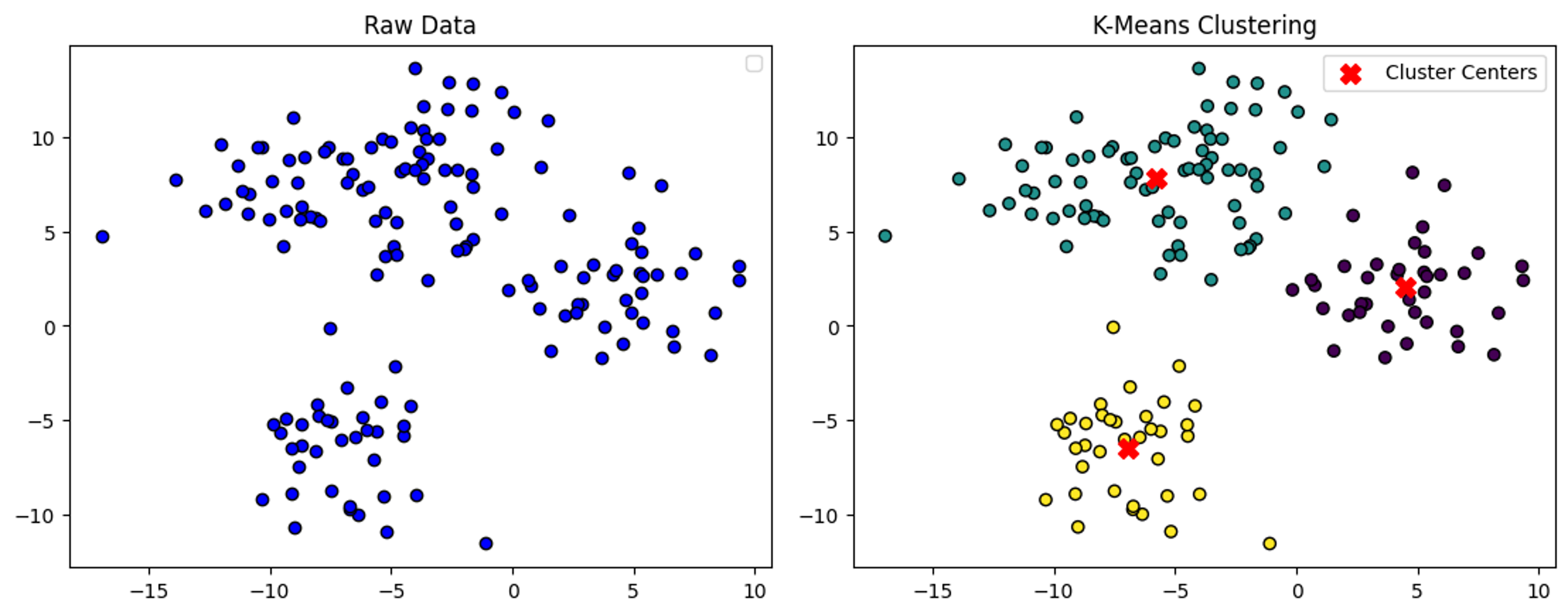}
        \caption{Example of Partition-Based Clustering Algorithm: the left side represents the original data, and the right side shows the resulting clusters after applying the K-Means clustering algorithm. Each data sample is classified into only one cluster. }
        \label{fig:Partition-Based}
    \end{figure}
    \item Hierarchical-Based Clustering (e.g. Figure~\ref{fig:Hierarchical-Based}): The sole concept of hierarchical clustering lies in the construction and analysis of a dendrogram. 
    A dendrogram is a tree-like structure that contains the relationship between all the data points in the system. There is no need to specify a predefined number of clusters. These algorithms can handle clusters of various shapes and sizes.
    Once the dendrogram is constructed, a horizontal cut through the structure defines individual clusters at the highest level in the system. Each resulting child branch below this cut represents a distinct cluster, assigning cluster membership to each data sample.
    However, hierarchical clustering needs intensive computation, especially for large datasets, and the time complexity is often higher than other clustering methods. 
    This structure is sensitive to noise and outliers, which can lead to sub-optimal clusters without the appropriate handle. 
    \begin{figure}[ht]
        \centering
        \includegraphics[width=1\linewidth]{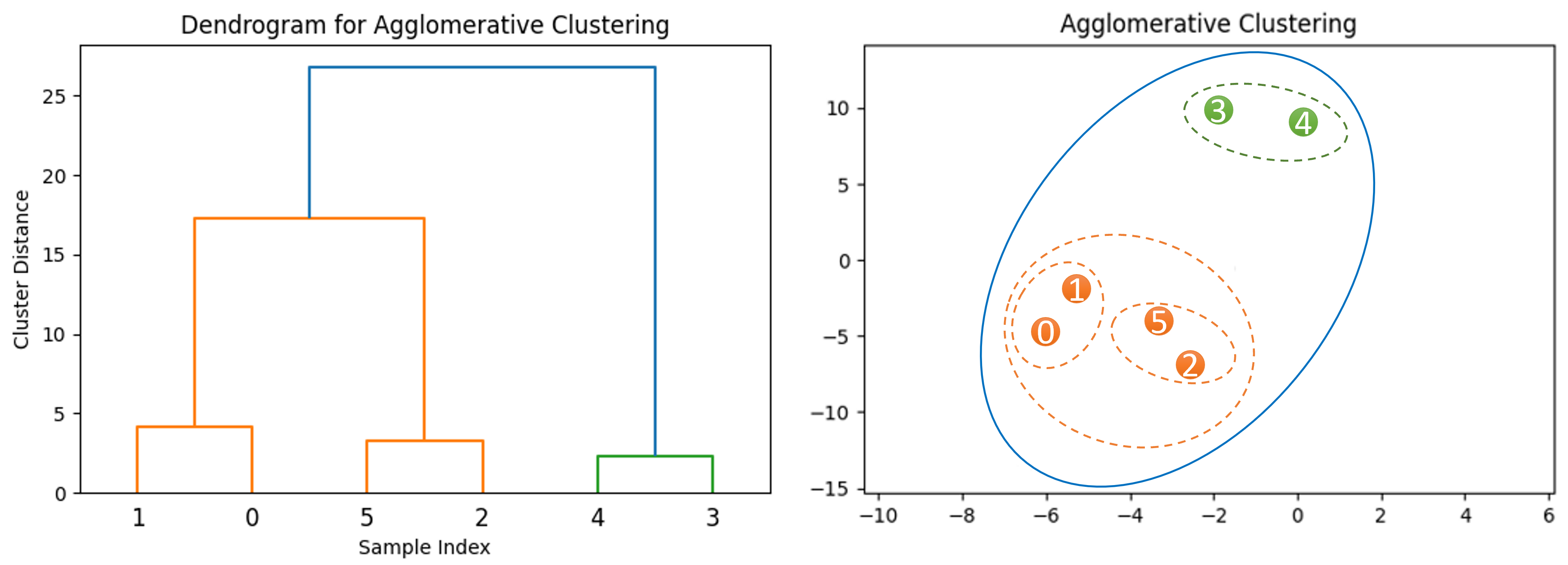}
        \caption{Schematic diagram of a hierarchical clustering algorithm:  the left side displays a dendrogram constructed from the relationships between samples in the dataset, while the right side illustrates the resulting clusters based on the dendrogram.}
        \label{fig:Hierarchical-Based}
    \end{figure}
    \item Density-Based Clustering (e.g. Figure~\ref{fig:Density-Based}): Density-based algorithms aim to identify clusters in high-dimensional regions of the feature space while also detecting noise points as outliers. These algorithms have high clustering efficiency, are sensitive to parameters, and are suitable for datasets of arbitrary shapes. In the case of uneven spatial data density, the quality of clustering results will decrease. Additionally, higher computing resources are required when processing large datasets.

    \begin{figure}[ht]
        \centering
        \includegraphics[width=1\linewidth]{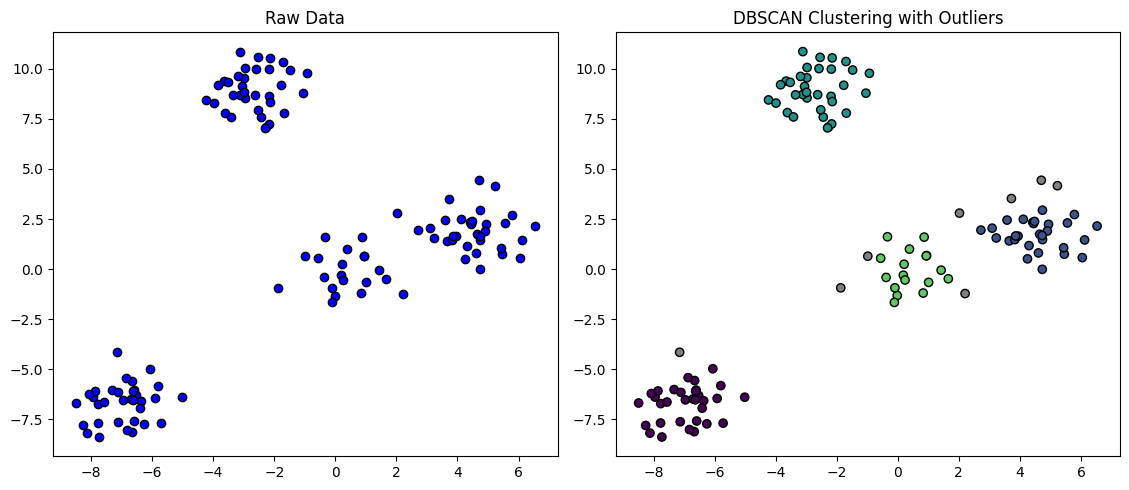}
        \caption{Schematic diagram of density-based clustering algorithm: the left side represents the original data, and the right side shows the clustering results after applying the algorithm. There are some outliers, represented by gray dots in the graph.}
        \label{fig:Density-Based}
    \end{figure}
     
    \item Grid-Based Clustering (e.g. Figure~\ref{fig:Grid_Based}): The fundamental principle of these clustering algorithms is to partition the initial data space into a grid structure of a predetermined size for clustering. 
    While exhibiting low time complexity, high scalability, and compatibility with parallel processing and incremental updates, these algorithms do come with considerations. The clustering outcomes prove sensitive to the grid size, where the pursuit of heightened calculation efficiency may come at the expense of cluster quality and overall clustering accuracy.
    \begin{figure}
        \centering
        \includegraphics[width=1\linewidth]{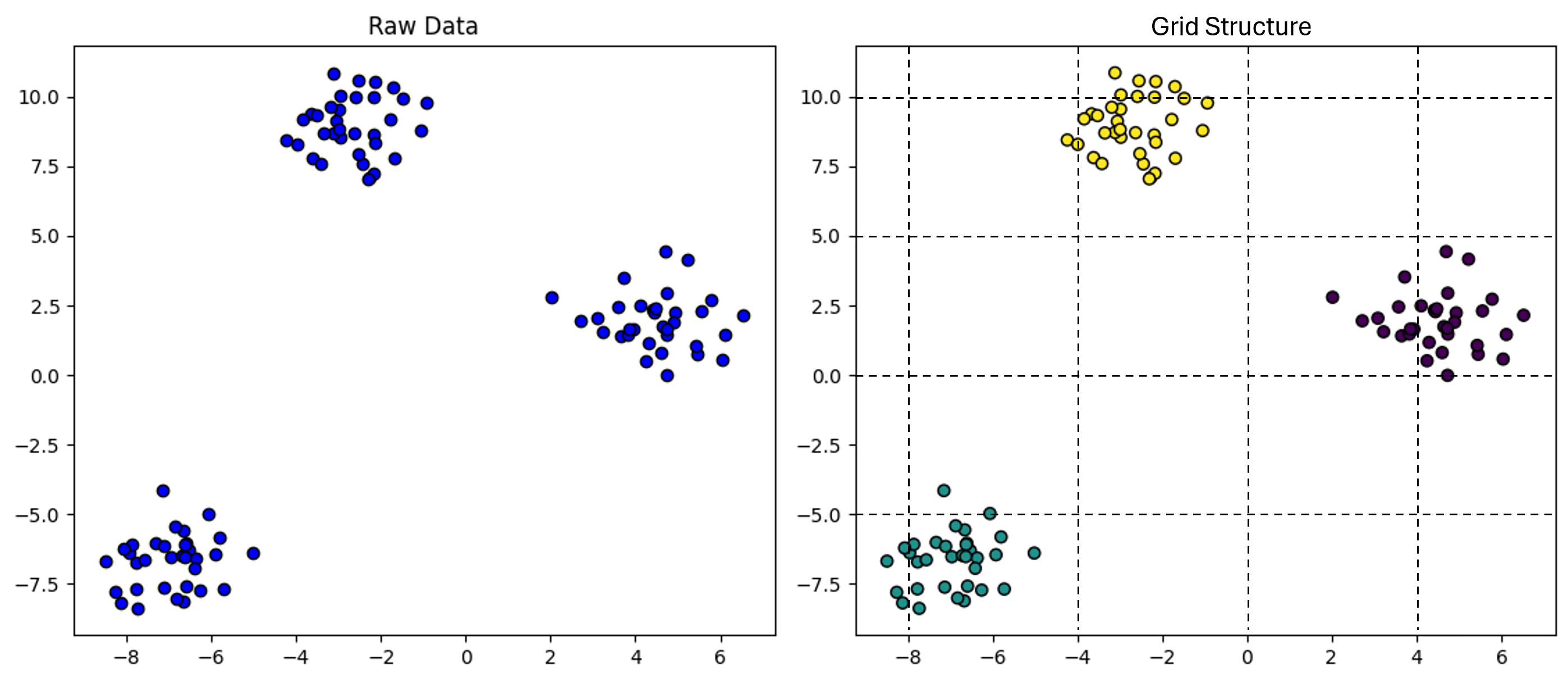}
        \caption{Schematic diagram of grid-based clustering algorithm: the left figure represents the original data, and the right is a schematic grid algorithm diagram. The grid might allow for arbitrary shapes or adaptations to suit the characteristics of the data better. The identification of clusters is typically done by defining rules or criteria based on the occupancy of grid cells.}
        \label{fig:Grid_Based}
    \end{figure}
    \item Model-Based Clustering (e.g. Figure~\ref{fig:Model_Based}): The basic idea is to select a particular model for each cluster and find the best fitting for that model. Model-based clustering algorithms presume data points are generated from a probabilistic model and seek to identify the most appropriate model to explain the data distribution. Diverse and well-developed models provide means to describe data adequately, and each model has its unique characteristics that may offer some notable benefits in some specific areas. However, overall, the time complexity of the models is relatively high, the premise is not entirely true, and the clustering result is dependent on the parameters of the models that are chosen.  
    \begin{figure}
        \centering
        \includegraphics[width=1\linewidth]{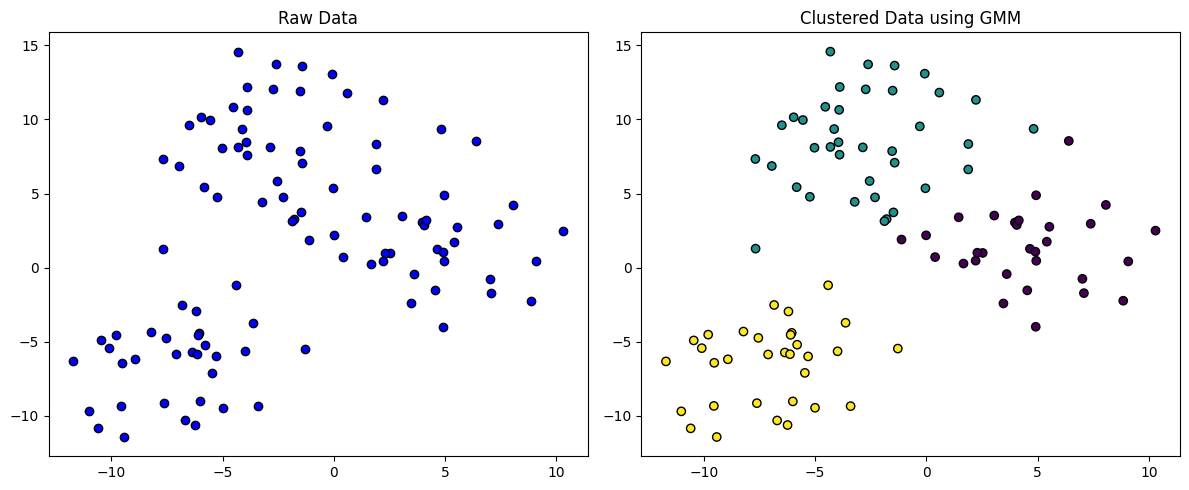}
        \caption{Schematic diagram of model-based clustering algorithm. The left side represents the original data, and the right side shows the clustering results after applying the Gaussian Mixture Model (GMM). Real-world clustering results may vary due to different model choices, parameter settings, and other factors.}
        \label{fig:Model_Based}
    \end{figure}
\end{itemize}

\begin{table*}[h!]
\caption{Algorithms classified by underlying principles and characteristics.}
\label{tab_Principles}
\begin{tabular}{p{1.8cm}|>{\raggedright\arraybackslash}p{3.2cm}|>{\raggedright\arraybackslash}p{3.2cm}|>{\raggedright\arraybackslash}p{4cm}}
\hline
\hline


Category & Benefits & Limitations & Algorithm Examples \\ \hline

Partition-Based Clustering & Efficient (low time complexity). Relatively simple to implement. & Sensitive to outliers. Drawn to local optima. Requires predefined number of clusters. & K-Means~\cite{Lloyd1982,Forgy1965}, K-medoids~\cite{kaufman1990partitioning,schubert2021fast,rdusseeun1987clustering}, PAM~\cite{kaufman1990partitioning},  CLARA~\cite{kaufman2009finding},  CLARANS~\cite{ng2002clarans}, FCM~\cite{bezdek1984fcm} \\ \hline

Hierarchical Clustering & Does not require a predefined number of clusters. Dendrogram provides heirarchical information on clusters, subclusters. & Inefficient (high time complexity). Dendrogram structure sensitive to noise and outliers.  & Agglomerative Clustering~\cite{johnson1967hierarchical}, Chameleon~\cite{karypis1999chameleon},  BIRCH~\cite{zhang1996birch}, CURE~\cite{guha1998cure}, Complete Linkage ~\cite{sneath1973numerical}, ROCK~\cite{guha2000rock}, Single Linkage Clustering~\cite{sneath1973numerical}, Divisive Clustering~\cite{savaresi2002cluster} \\ \hline

Density-Based Clustering & High clustering efficiency. Suitable for arbitrary cluster shapes. & Decreased clustering quality for uneven spatial data density. Inefficient for large datasets.  & DBSCAN~\cite{ester1996density}, DENCLUE~\cite{hinneburg1998efficient}, HDBSCAN~\cite{campello2013density}, Mean-Shift Clustering~\cite{comaniciu2002mean}, OPTICS~\cite{ankerst1999optics} \\ \hline

Grid-Based Clustering & Efficient (low time complexity). Suitable for arbitrary cluster shapes.  & Clusters sensitive to grid size. & WaveCluster, CLIQUE~\cite{agrawal1998automatic}, STDBSCAN~\cite{birant2007st},  Grid-DBSCAN~\cite{ester1996density}, STING~\cite{wang1997sting} \\ \hline

Model-Based Clustering & Allows for non-spherical clusters. Produces uncertainty of cluster membership. Does not require a predefined number of clusters. & Relatively inefficient (high time complexity). & GMM~\cite{rasmussen1999infinite}, Hidden Markov Models (HMM)~\cite{rabiner1986introduction}, Latent Dirichlet Allocation (LDA)~\cite{blei2003latent}, Expectation-Maximization Clustering (EM)~\cite{dempster1977maximum} \\ \hline
\end{tabular}
\end{table*}

\subsection{Algorithm Classification by Data Point Assignment to Clusters: Single vs. Multi}
Clustering algorithms can be broadly categorized based on how they assign data points to clusters. This leads to two primary classifications: hard clustering, where each data point is exclusively associated with a single cluster, and soft clustering, where data points can be simultaneously associated with multiple clusters. Below are some classic algorithms under each category.
\begin{itemize}
    \item Hard clustering: K-Means, hierarchical clustering (agglomerative), DBSCAN, OPTICS, K-Medoids, Spectral Clustering (K-way Normalized Cut)~\cite{dhillon2004kernel}, BIRCH.
    \item Soft clustering: Fuzzy C-Means (FCM), Gaussian Mixture Model (GMM), Possibilistic C-Means (PCM)~\cite{krishnapuram1996possibilistic}, Possibilistic Fuzzy C-Means (PFCM)~\cite{pal2005possibilistic}, Kernelized Fuzzy C-Means (KFCM)~\cite{zhang2004novel}, Hierarchical Fuzzy Clustering (HFC)~\cite{di1994integrated}
\end{itemize}

\subsection{Algorithm Classified by Dataset Capacity}
Algorithms differ in their ability to handle dataset size, and in this section we classify all algorithms into three categories based on their processing capabilities. Of course, defining what constitutes a small, medium, or large dataset can be somewhat subjective, depending on the the context of your specific task.
Based on an overview of the open-source library for Python~\footnote{\url{https://scikit-learn.org/stable/modules/clustering.html\#overview-of-clustering-methods}} and existing work, we categorize algorithms into three groups according to their processing capabilities, as illustrated in Table~\ref{tab_DatasetSize}.

\textbf{Small Dataset:}
Typically, a small dataset might contain a few hundred to a few thousand instances.
It is a scale where the entire dataset can be easily loaded into memory and processed without significant computational resources.

\textbf{Medium Dataset:}
A medium-sized dataset could range from a few thousand to tens of thousands of instances.
It might require more sophisticated algorithms and computational resources compared to a small dataset, but it is still manageable.

\textbf{Large Dataset:}
Large datasets typically contain hundreds of thousands to millions (or more) instances.
Handling such datasets often requires specialized algorithms, distributed computing, or parallel processing due to the sheer volume of data.

\begin{table*}[]
\caption{Algorithms classified by dataset capacity.}
\label{tab_DatasetSize}
\begin{tabular}{p{2.2cm}|p{4.5cm}|p{5cm}}
\hline
\hline
Dataset & Algorithm Examples & Comments \\ \hline
Small (up to a few thousand instances) & K-Means, DBSCAN, Hierarchical clustering &  These algorithms are often efficient and can run on standard machines thereby avoiding the need for extensive computational resources. \\ \hline
Medium (thousands to hundreds of thousands) & Optimized K-Means, GMM, Optimized Agglomerative clustering, Mean-Shift clustering  & The algorithms may utilize  more sophisticated techniques to deal with the increased size of the dataset. \\ \hline
Large~\cite{shirkhorshidi2014big,kurasova2014strategies} (hundreds of thousands or more) & Mini-Batch K-Means, BIRCH, DBSCAN (optimized for large datasets), Hierarchical clustering (optimized for large datasets) &   These algorithms are tailored for parallel processing to efficiently tackle computational challenges in handling large  datasets. \\ \hline
\end{tabular}
\end{table*}

\subsection{Algorithm Classification by Predefined Cluster Numbers}
\label{sec_PredifinedClusterNumbers}
Not all clustering algorithms require a predefined number of clusters. There are two main types of clustering algorithms:

\textbf{Hard clustering:}
Algorithms in this category, such as K-Means, require the user to specify the number of clusters beforehand.

\textbf{Soft clustering or hierarchical clustering:}
It is not necessary to specify the number of clusters in advance when using algorithms like hierarchical clustering or GMMs. Clustering can be done at different levels of granularity or number of clusters, and you can choose which to use post-hoc.

The biggest challenge in hard clustering is how to determine the optimal number of clusters.
Clustering results can be interpreted differently depending on the number of clusters used, the granularity of patterns discovered in the data, and the practicality of the solution.
As a crucial step in the clustering process, choosing the right number of clusters also influences the insights derived from the analysis and the usefulness of the clustered groups in real-life situations.
There are some methods that can be used to determine the optimal number of clusters. 
\begin{itemize}
    \item Elbow Method~\footnote{\url{https://en.wikipedia.org/wiki/Elbow_method_(clustering)}}: Run the clustering algorithm for a range of cluster numbers and plot the within-cluster sum of squares (inertia) against the number of clusters.
    Look for an ``elbow" in the plot, where the rate of decrease in inertia slows down. The point where this occurs is often considered the optimal number of clusters. There are several publications that employ the elbow method to determine the optimal number of clusters~\cite{syakur2018integration,bholowalia2014ebk,marutho2018determination}.
    \item Silhouette Score~\footnote{\url{https://en.wikipedia.org/wiki/Silhouette_(clustering)}}:  Refers to a method of interpretation and validation of consistency within clusters of data.
    The silhouette score measures how similar an object is to its own cluster compared to other clusters. Choose the number of clusters that maximizes the silhouette score.
    There are several publications that employ the silhouette score to evaluate the optimal number of clusters~\cite{shutaywi2021silhouette,ogbuabor2018clustering,shahapure2020cluster}
    \item Gap Statistics~\cite{tibshirani2001estimating}: Compare the clustering performance on your dataset with the performance on a random dataset (or with fewer clusters).
    Optimal clusters should have a larger gap in performance compared to random clustering. Here are some publications that employ gap statistic to determine the number of clusters~\cite{yan2007determining,el2019optimized,mohajer2011comparison}
    \item Dendrogram~\cite{calinski2014dendrogram} in Hierarchical Clustering:
    If the task applies to a hierarchical clustering algorithm, visualize the dendrogram and look for a level where cutting it results in a reasonable number of distinct clusters. There are some publications using dendrogram for clustering~\cite{langfelder2008defining,nielsen2016hierarchical}
\end{itemize}

\subsection{Algorithm Classification by Area of Application}
In this section, we will compare various key application areas and the clustering algorithms predominantly utilized in each domain, shown in Table~\ref{tab_areas}. In fields like data mining and information retrieval, algorithms like K-Means and DBSCAN are frequently employed for their efficiency in handling large datasets. In contrast, areas like image analysis and bioinformatics often rely on algorithms like Spectral Clustering and Hierarchical Clustering, which are better suited for these fields' intricate patterns and structures. It is worth noting that new algorithms like AutoClass~\cite{li2022universal} and Superpixel~\cite{asantemensah2023image} work well in specific complicated scenarios.


\begin{table*}[]
\caption{Areas of application of clustering algorithms.}
\label{tab_areas}
\begin{tabular}{p{3.5cm}|p{8.5cm}}
\hline
\hline
Area of Application & Algorithm Examples \\ \hline
Data Mining & K-Means, DBSCAN \\ \hline
Image Analysis & Spectral Clustering, Mean Shift \\ \hline
Pattern Recognition & Hierarchical Clustering, K-Means \\ \hline
Information Retrieval & Hierarchical Clustering, K-Means, DBSCAN \\ \hline
Bioinformatics & Hierarchical Clustering, Spectral Clustering \\ \hline
Image reconstruction & K-Means, Superpixel, Expectation Maximization \\ \hline
Network Analysis & Hierarchical Clustering, K-Means, DBSCAN, AutoClass \\ \hline

\end{tabular}
\end{table*}







\section{Evaluation Metrics}\label{sec_Evaluation}
\begin{figure}
    \centering
    \includegraphics[width=0.7\linewidth]{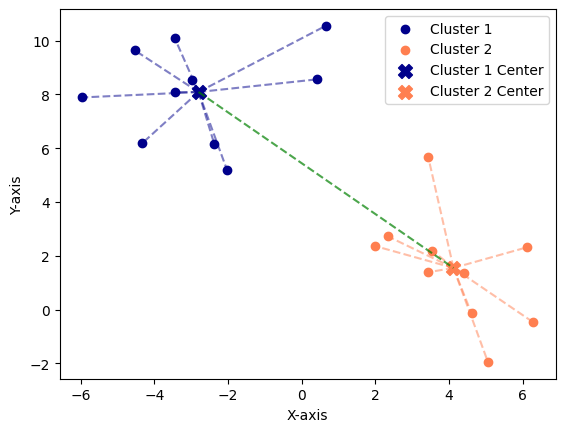}
    \caption{Example of K-Means clustering with two clusters, illustrating two different types of distance between data points.}
    \label{fig:PointDis}
\end{figure}

While most datasets lack ground truth labels for clustering algorithms, methods exist for evaluating clustering quality. Evaluation metrics, crucial for guiding the development, selection, and optimization of clustering algorithms, make the process more systematic and informed.
Data samples transform into vectors in a high-dimensional space during the clustering process. 
The distances between these vectors intricately reflect the overall similarity, incorporating all relevant features within the data samples.
Therefore, ``distances between points" are pivotal in forming clusters and serve as a common standard for evaluating clustering performance. This concept refers to numerical measures of dissimilarity or similarity between individual data points, often calculated based on data features. Typical metrics include Euclidean distance, Manhattan distance, or other similarity measures like cosine similarity for text data.
Figure~\ref{fig:PointDis} illustrates an example of K-Means clustering, showcasing the distances between data points. Two types of distances are depicted: one represents the distance between data samples within the cluster (blue and orange dashed lines), while the other indicates the distance between different cluster centers (green dashed line). The distance between data points within a cluster is considerably smaller than between data points and the centers of other clusters.

By emphasizing the significance of evaluation metrics and the role of distances between points, the process of assessing and improving clustering algorithms becomes more robust and meaningful.
Evaluation metrics are categorized into internal and external types.
Internal metrics assess the quality of clusters based solely on inherent data information, focusing on factors like the compactness (cohesion) of data samples within a cluster and the distinctiveness (separation ) between clusters.
Examples include the Silhouette Coefficient, Davies-Bouldin Index, Dunn's Index, and Inertia (for K-Means). External metrics require true class labels, evaluating alignment with known labels using precision, recall, and mutual information. Such as the Adjusted Rand Index (ARI), Normalized Mutual Information (NMI), Fowlkes-Mallows Index, Precision, Recall, and F1 Score. Internal metrics focus on intrinsic properties, while external metrics rely on ground truth to evaluate algorithm accuracy. The choice depends on data availability and clustering goals.

\subsection{Internal Evaluation Metrics}
Internal evaluation metrics for clustering algorithms can evaluate the quality of clusters based solely on the data's intrinsic characteristics and the clustering algorithm's results, without using any external information, such as ground truth labels. 
Various aspects of clustering quality can be measured quantitatively with these metrics, such as compactness, separation, and variance. When choosing and interpreting these metrics, it is essential to consider the specific characteristics of the data and the goals of the clustering task.

\subsubsection{Silhouette Score} The Silhouette Score is a widely used internal evaluation metric for assessing the quality of clusters produced by unsupervised clustering algorithms. It measures how well-separated clusters are and provides an indication of the appropriateness of the clustering solution. The Silhouette Score for a single data point is calculated as the difference between the average distance from the data point to other points in the same cluster (a) and the average distance from the data point to points in the nearest neighboring cluster (b), divided by the maximum of a and b.
For a data point $i$, the formula is:
\begin{equation}
    S(i)=\frac{b(i)-a(i)}{max\left\{ {a(i),b(i)}\right\}}
\end{equation}

The Silhouette Score ranges from -1 to 1, with higher values indicating better-defined clusters.
The Score close to 1 indicates that the data point is well matched to its own cluster and poorly matched to neighboring clusters, suggesting a good clustering.
A Silhouette Score around 0 indicates overlapping clusters.
A Silhouette Score close to -1 indicates that the data point may be assigned to the wrong cluster.
The overall Silhouette Score for a clustering solution is the average of the Silhouette Scores for all data points.

\subsubsection{Davies-Bouldin Index} The Davies-Bouldin Index is a metric used for evaluating the quality of clustering solutions. It measures the compactness and separation between clusters, aiming to find clusters that are well-separated from each other.
The formula of Davies-Bouldin Index is:

\begin{equation}
    DBI=\frac{1}{n_{c}}\sum_{i=1}^{n_{c}}max_{i\neq j}( \frac{{avg_{radius_{i}}+avg_{radius_{j}}}}{distance(c_{i},c_{j})})
\end{equation}

where: $n_{c}$ is the number of clusters, $c_i$ and $c_j$ are the centroids of clusters $i$ and $j$, ${avg_{-}radius}_{i}$ and ${avg_{-}radius}_{j}$ are the average distances from the centroid to the points in custer $i$ and $j$, distance$(c_i,c_j)$ is the distance between the centroids of clusters $i$ and $j$.

The Davies-Bouldin Index is based on the Euclidean distance, which assumes clusters have a spherical shape and may not perform well when dealing with clusters of irregular shapes.
Also, the formula assumes equal-sized clusters, so it might not be suitable for datasets with imbalanced cluster sizes.
It may not perform well if the true number of clusters is unknown or if the clustering algorithm produces a different number of clusters.
Despite these limitations, the Davies-Bouldin Index can be a useful tool when the assumptions align with the characteristics of the data and the goals of the clustering analysis.

\subsubsection{Dunn's Index} The Dunn's Index focuses on the compactness and separation of clusters. It aims to find a balance between minimizing the diameter (maximum distance between points) within a cluster and maximizing the distance between cluster centroids. The index is defined as the ratio of the minimum inter-cluster distance to the maximum intra-cluster diameter.

\begin{equation}
    DBI=\frac{1}{n}\sum_{i=1}^n max_{j\neq i}(\frac{avg_{-}intra_{-}distance(C_i)+avg_{-}intra_{-}{distance}(C_{j})}{distance(c_{i},c_{j})}
\end{equation}

while $n$ is the number of clusters, $C_{i}$ and $C_{j}$ are clusters, $avg_{-}intra_{-}distance(C_{i})$ is the average distance within cluster $C_{i}$, and $distance(c_{i},c_{j})$ is the distance between cluster centers $c_{i}$ and$c_{j}$.

Dunn's Index is sensitive to outliers, and highly dependent on the distance metric.
It assumes that clusters are spherical, so if the clusters have non-spherical shapes, it may not accurately reflect the true separation between clusters.
Dunn's Index produces a numeric result, but its interpretation as ``good" or ``bad" is subjective and context-dependent in clustering problems.
Despite these shortcomings, Dunn's Index can still be a valuable tool when used judiciously and in conjunction with other evaluation metrics.

The above three are commonly used evaluation metrics, there are other less popular evaluation metrics, such as Calinski-Harabasz Index~\cite{calinski1974dendrite} which evaluates the ratio of the between-cluster variance to the within-cluster variance, the higher values the better-defined clusters. Inertia (Within-Cluster Sum of Squares)~\cite{macqueen1967some} which measures the sum of squared distances between data points and their cluster's centroid. Lower inertia suggests denser, more compact clusters. Gap Statistics~\cite{tibshirani2001estimating} which compares the clustering quality of the dataset to that of a reference random dataset. A larger gap indicates better clustering. CH Index (Cophenetic Correlation Coefficient)~\cite{sneath1973numerical} which measures the correlation between the cophenetic distances in the dendrogram and the original distances. Higher CH Index suggests better clustering.

\subsection{External Evaluation Metrics (when ground truth is available)}
\subsubsection{Adjusted Rand Index (ARI)}
The Adjusted Rand Index (ARI) is a widely used metric in cluster analysis and machine learning for evaluating the similarity between two clustering solutions. It measures the agreement between the true class labels and the labels assigned by a clustering algorithm while correcting for chance.
The formula of ARI is:

\begin{equation}
    ARI=\frac{\text{RI}-\text{Expected RI}}{\text{Max(RI)}-\text{Expected RI}}
\end{equation}

Where $RI$ is the Rand Index, which measures the proportion of agreements (both in the same cluster or both in different clusters) between the true and predicted clusterings.
$Expected_{RI}$ is the expected Rand Index under the assumption of random clustering. It represents the expected value of $RI$ when clustering is performed randomly.
The $max(RI)$ term in the denominator represents the maximum possible Rand Index, which normalizes the ARI to the range [-1, 1].

The limitations of ARI are as follows: Sensitivity to Imbalanced Cluster Sizes: ARI can be sensitive to imbalanced cluster sizes. If there is a significant difference in the number of samples in different clusters, ARI may be biased towards the larger clusters. Dependence on the Number of Clusters: ARI assumes knowledge of the true number of clusters. If the true number of clusters is unknown or if the clustering algorithm produces a different number of clusters, ARI might not provide an accurate evaluation.
Random Clustering Assumption: ARI's correction for chance assumes that cluster assignments are made randomly. In some cases, especially with certain clustering algorithms or data types, the assumption of random clustering might not hold.
Limited to Pairwise Comparisons: ARI is designed for pairwise cluster comparison and does not provide information on the overall structure of multiple clusters. It may not capture more complex relationships in the data.
Dependency on Ground Truth: ARI requires knowledge of true class labels, which may not be available in unsupervised learning scenarios. In such cases, alternative evaluation metrics may be needed. 
Despite these limitations, ARI remains a widely used and interpretable metric for clustering evaluation. It is important to consider these shortcomings in the context of your specific clustering task and choose evaluation metrics accordingly.

\subsubsection{Normalized Mutual Information (NMI)} Normalized Mutual Information provides a quantitative measure of the agreement between the true class labels and the labels assigned by a clustering algorithm, taking into account both precision and recall, offering a balanced perspective on the quality of the clustering solution.

The formula of NMI is:
\begin{equation}
    NMI=\frac{2 \quad I(Y;C)}{H(Y)+H(C)}
\end{equation}

where $Y$ is the set of true class labels, $C$ is the set of cluster labels assigned by the algorithm, $I(Y;C)$ is the mutual information between $Y$ and $C$, and $H(Y)$ and $H(C)$ are the entropies of $Y$ and $C$ respectively.
NMI ranges from 0 to 1, where 0 indicates no mutual information, and 1 implies perfect agreement between the true and predicted labels.
A higher NMI suggests a better clustering solution in terms of capturing the underlying class structure.
NMI is a commonly used metric in most of clustering solution, as it accounts for both homogeneity and completeness in clustering evaluation and normalization helps in comparing NMI across datasets of different sizes.
The limitation of NMI is it assumes that each cluster corresponds to a single class, which may not always be the case in real-world data.

\section{Discussion}\label{sec12}
In recent years, there has been a shift in the focus of clustering algorithm research from solely improving the underlying algorithms to more targeted applications in specific fields. This shift is driven by the increasing recognition of the diverse and complex data challenges in various domains. 
Researchers are now actively exploring how clustering algorithms can be effectively applied and adapted to address the unique requirements of fields such as bioinformatics~\cite{karim2021deep,higham2007spectral,olman2008parallel}, healthcare~\cite{haraty2015enhanced,ogbuabor2018clustering,delias2015supporting}, natural language processing~\cite{yang2019discovering,yin2021representation}, image and video processing~\cite{Cao2023,dhanachandra2015image}, social network analysis~\cite{jose2019detecting,zhao2011new,8056956}, cybersecurity~\cite{8268746,8526232,kolini2017clustering,landauer2020system}, and anomaly detection~\cite{agrawal1998automatic,syarif2012unsupervised,markovitz2020graph,aggarwal2012survey}. 
The research community has concentrated its efforts on customizing clustering solutions to align with particular application contexts, thereby facilitating significant progress in domain-specific applications of clustering methodologies.
We observed that the COVID-19 pandemic has led to a significant increase in the use of clustering algorithms in medical imaging and healthcare from 2021 to 2022. 
Amidst the swift evolution of deep learning technology, a discernible trend has emerged in the advancement of clustering algorithms.
Specifically, there is an increasingly obvious trend to incorporate deep learning technologies, such as neural networks, into clustering algorithms to improve their performance~\cite{aljalbout2018clustering,shaham2018spectralnet,pmlr-v119-bianchi20a}, particularly in processing high-dimensional and complex data.
K-Means is one of the oldest and most well-known clustering algorithms, having achieved popularity over the years for its simplicity and effectiveness in partitioning data into clusters based on similarity. 
Despite its age, K-Means continues to be widely used in various applications and is a familiar presence in clustering technical publications. It often serves as a baseline method for comparison with proposed approaches, and ongoing efforts are made to enhance performance~\cite{9072123,yu2018two,fard2020deep,ran2021novel}.
Another noteworthy observation is the increasing popularity of hybrid clustering methods~\cite{7283582,you2023applying}. These methods combine different clustering algorithms or integrate clustering with other machine learning techniques. These approaches aim to leverage the strengths of multiple methods for enhanced performance.

Currently, the primary challenge confronting clustering algorithms revolves around determining the optimal number of clusters. Groups generated using existing techniques often exhibit ambiguous clusters—instances wherein data is grouped into unspecified categories for reasons unknown. If only one group is present, it can be classified as an outlier, but occasionally, multiple groups remain unidentified.
This highlights a potential discrepancy between the algorithm's grouping of data and the subjective human interpretation of those groups.
In recent academic literature, several different clustering methods are frequently used, each with its own strengths and applications. Therefore, the choice of a clustering method is highly task-dependent. There is no single method universally outperforming others across all types of data and applications.
Our review classifies algorithms from multiple perspectives and can assist users in choosing the appropriate clustering algorithm for a given application.
\section*{Acknowledgement}
This research was supported by the Australian Government through the Australian Research Council's Industrial Transformation Training Centre for Information Resilience (CIRES) project number IC200100022.

\noindent We would like to express our sincere gratitude to Junliang Yu, Luhan Cheng, Nakul Nambiar, Yunzhong Zhang, Shuyi Shen, and Zhuochen Wu for their valuable contributions and insightful feedback during the development of this work.
 
\section*{Declarations}
\textbf{Conflict of interest} The authors declare that they have no known competing financial interests or personal relationships that could have appeared to influence the work reported in this paper.

\bibliography{sn-bibliography}

\end{document}